\documentclass[10pt,conference,a4paper]{IEEEtran}

\usepackage{amsmath}
\usepackage{amssymb}
\usepackage{url}
\usepackage{ogonek}
\usepackage{graphicx}

\begin{document}
%
\title{Classification of COPD with \\ Multiple Instance Learning}
%
%
%

\author{\authorblockN{Veronika Cheplygina\authorrefmark{1}, Lauge S{\o}rensen\authorrefmark{2}
, David M. J. Tax\authorrefmark{1},\\Jesper Holst Pedersen\authorrefmark{3}, Marco Loog\authorrefmark{1}\authorrefmark{2} and Marleen de Bruijne\authorrefmark{2}\authorrefmark{4}}

\authorblockA{\authorrefmark{1}Pattern Recognition Laboratory, Delft University of Technology, Delft, The Netherlands\\
}
\authorblockA{\authorrefmark{2}The Image Group, Department of Computer Science, University of Copenhagen, Copenhagen, Denmark\\
}

\authorblockA{\authorrefmark{3}Department of Thoracic Surgery, Rigshospitalet, University of Copenhagen, Copenhagen, Denmark\\
}

\authorblockA{\authorrefmark{4}Biomedical Imaging Group Rotterdam, Erasmus MC, Rotterdam, The Netherlands\\
\emph{Email: v.cheplygina@tue.nl, \{d.m.j.tax, m.loog\}@tudelft.nl, \{lauges,marleen\}@diku.dk}
}
}


%
%

%

\maketitle

\begin{abstract}
Chronic obstructive pulmonary disease (COPD) is a lung disease where early detection benefits the survival rate. COPD can be quantified by classifying patches of computed tomography images, and combining patch labels into an overall diagnosis for the image. As labeled patches are often not available, image labels are propagated to the patches, incorrectly labeling healthy patches in COPD patients as being affected by the disease. We approach quantification of COPD from lung images as a multiple instance learning (MIL) problem, which is more suitable for such weakly labeled data. We investigate various MIL assumptions in the context of COPD and show that although a concept region with COPD-related disease patterns is present, considering the whole distribution of lung tissue patches improves the performance. The best method is based on averaging instances and obtains an AUC of 0.742, which is higher than the previously reported best of 0.713 on the same dataset. Using the full training set further increases performance to 0.776, which is significantly higher (DeLong test) than previous results.

\end{abstract}

\begin{IEEEkeywords}
Computer aided diagnosis, chronic obstructive pulmonary disease, supervised learning, multiple instance learning
\end{IEEEkeywords}

%

\section{Introduction}\label{sec:intro}

Chronic obstructive pulmonary disease (COPD) is a disease of the lungs that is caused, among others, by smoking and air pollution. COPD is characterized by chronic inflammation of the lung airways, and degradation of lung tissue, called emphysema, both of which result in airflow limitation~\cite{calverley2000copdearly,rabe2007global}. The disease progresses in several stages and can eventually lead to death, however, detecting the disease at an early stage can increase the survival rate~\cite{pauwels2012global}.

Due to limitations of traditional spirometry and visual assessment of computed tomography (CT) scans, texture classification was proposed to quantify COPD~\cite{park2008texture,sorensen2012texture,srensen2010quantitative,uppaluri1997quantification,mendoza2012emphysema}. One approach is to classify patches of lung tissue, or regions of interest (ROIs) in the image, and combine the classifications into an overall probability for COPD~\cite{park2008texture,srensen2010quantitative}. However, these supervised approaches require manually annotated ROIs, which are difficult and costly to obtain.

An alternative is to use weakly labeled medical images, i.e., where only a global image label is provided, for training an image classifier. In the absence of labeled ROIs, the image label can be propagated to its ROIs, and an ROI classifier can be trained as usual~\cite{sorensen2012texture}. We call this straightforward approach SimpleMIL. However, this disregards the fact that in scans of patients with COPD, only a subset of the ROIs may be affected, while signs of COPD may be already apparent in some regions for subjects not yet diagnosed with the disease. This increases the label noise for the ROI classifier.

A technique which can handle learning with such weakly labeled data is called multiple instance learning (MIL)~\cite{dietterich1997solving,maron1998framework}. The goal is to build a classifier for a collection, or bag, of feature vectors, also referred to as instances. Often it is assumed that a bag is positive if and only if at least one of its instances is positive. A further assumption is that positive instances are found in a region of the feature space called the concept. For COPD, the concept could be a part of the feature space, containing ROIs that are typical for, for example, emphysema. In this scenario, as soon as a CT scan contains such an ROI, the whole image is diagnosed as COPD.

MIL methods can be broadly divided into two categories: instance-based and bag-based. Instance-based methods use the constraints posed by the bag labels and the MIL assumptions to build an instance classifier, and combine instance classifications to classify bags~\cite{maron1998framework,zhang2001dd,andrews2002support,viola2006multiple}. On the other hand, bag-based methods aim to classify bags directly, often by defining kernels~\cite{gartner2002multi} or dissimilarities~\cite{tax2011bag,zhang2009multi2} between bags.

Every MIL classifier makes explicit or implicit assumptions about the data. Instance-based classifiers typically rely on the assumption that there is a concept, and that positive bags contain instances from this concept. Therefore, only concept instances are important for determining the bag label. Bag-based classifiers assume that bags from the same class are similar, and the similarity definition further specifies this assumption. In most definitions, all of the bag's instances are involved in defining the bag similarity, therefore the whole distribution of instances is important for the bag label. In~\cite{cheplygina2012does} we have shown that many well-known MIL problems fall into these two categories (concept and distribution) and that this property determines how many MIL methods perform on the data.

Detection of COPD using lung texture has been tackled by classifying patches and combining their outputs, an approach we call SimpleMIL, in~\cite{sorensen2012texture}. A more specialized MIL method, applied to this problem, is a dissimilarity-based approach in~\cite{sorensen2010image}, and it shows promising results. Other dissimilarity or kernel-based approaches, which focus on the airways rather than lung texture, have also been successful for COPD classification~\cite{sorensen2011dissimilarity,feragen2013geometric}. In this work we investigate a broader range of MIL methods for classification of the lung texture in COPD. We examine which assumptions, commonly used for the instance-based and bag-based methods, are more suitable for this problem, and demonstrate state-of-the-art results on a COPD dataset from the Danish Lung Cancer Screening Trial~\cite{pedersen2009danish}.

\section{Multiple Instance Learning}\label{sec:mil}

In multiple instance learning (MIL), an object is represented by a bag $B_i  = \{\mathbf{x}_{ik}| k=1,...,n_i\} \subset \mathbb{R}^d$ of $n_i$ instances, where the $k$-th instance is described by a $d$-dimensional feature vector $\mathbf{x}_{ik}$. The training set $\mathcal{X}_{tr} = \{(B_i, y_i) | i=1,...N\}$ consists of positive ($y_i = +1$) and negative ($y_i = -1$) bags. One way to deal with this type of input is to propagate the bag labels to the instances, and building an instance classifier.  A bag label is obtained by classifying that bag's instances, and combining the instance classifications, for example by fusing the posterior probabilities~\cite{loog2004static}. The noisy-or rule,

\begin{equation}\label{eq:noisyor}
\frac{p(y=1|B_i)}{p(y=-1|B_i)} = \frac{1 - \prod_{k=1}^{n_i} (1-p(z_{ik}=1|\mathbf{x}_{ik}))}{\prod_{k=1}^{n_i} p(z_{ik}=-1|\mathbf{x}_{ik})}
\end{equation}

reflects the standard assumption that a bag is positive if and only if at least one of the instances is positive. On the other hand, the average rule,

\begin{equation}\label{eq:average}
\frac{p(y=1|B_i)}{p(y=-1|B_i)} = \frac{1}{n_i} \sum_{k=1}^{n_i} \frac{ p(z_{ik}=1|\mathbf{x}_{ik})} {p(z_{ik}=-1|\mathbf{x}_{ik})}
\end{equation}

assumes that all instances contribute to the bag label. This fusion rule has been used in~\cite{sorensen2012texture}, by classifying ROIs with a nearest neighbor classifier, and combining the outputs to classify the entire image. We refer to this strategy as SimpleMIL in the experiments.

The standard assumption for MIL is that there are hidden instance labels $z_{ik}$ which relate to the bag labels as follows: a bag is positive if and only if it contains at least one positive, or \emph{concept} instance~\cite{dietterich1997solving}. The strategy of earlier MIL approaches was to model the concept: a region in feature space which contains at least one instance from each positive bag, but no instances from negative bags. Diverse density~\cite{maron1998framework} (DD) has been proposed to measure this property. For a given point $t$ in the feature space, $DD(t)$ measures the ratio between the number of positive bags which have instances near $t$, and the sum of distances of the negative instances to $t$. The point where DD is maximized, $t^*$ therefore corresponds to the target concept. Instances can be classified using their distance to $t^*$. However, the optimization problem suffers from local optima and, for the original DD algorithm, several restarts of the algorithm are needed. Therefore, an expectation-maximization version of this algorithm EM-DD~\cite{zhang2001dd} has been proposed. EM-DD has shown to perform well on a range of MIL problems, but is also very computationally intensive.

Several regular supervised classifiers have been extended to work in the MIL setting. One example is mi-SVM~\cite{andrews2002support}, an extension of support vector machines which attempts to find hidden labels of the instances under constraints, such as~(\ref{eq:noisyor}) or~(\ref{eq:average}), posed by the bag labels. Another example is MILBoost~\cite{viola2006multiple}, where the instances are reweighed in each of the boosting rounds. The instance weights indicate how informative the instances are in predicting the bag labels.

It has been recognized that the standard assumption might be too strict for certain types of MIL problems. Therefore, relaxed assumptions have emerged~\cite{weidmann2003two}, where a fraction or a particular number of positive instances are needed to satisfy a concept, and where multiple concept regions are considered. In the case of COPD, this would correspond to the presence of a certain fraction of ROIs containing affected tissue, and/or different types of disease patterns. However, if the number of concepts and the fraction of positives per concept, are not given in advance, these extra parameters also need to be set using the training data, further increasing the risk of overtraining.

Therefore, methods which compare bags without explicitly relying on the standard, or relaxed assumptions, have been proposed. Such methods include Citation-$k$NN~\cite{wang2000solving}, and bag kernels~\cite{gartner2002multi}. Citation-$k$NN uses the Hausdorff distance between bags. For classification, both the $k_R$ ``referencing'' nearest neighbors of a bag $B$, and the $k_C$ ``citing'' neighbors (bags for which $B$ is nearest neighbor) are taken into account.
In~\cite{gartner2002multi}, a bag kernel is defined either as a sum of the instance kernels, or as a standard (linear or radial basis) kernel on a summarized representation of the bag. This summary is created by, for each feature, averaging the bag's instances (which we refer to by mean-inst), or using both the minimum and the maximum instance values (which we refer to by extremes). In all cases, the way a kernel is defined affects which (implicit) assumptions are made about the problem. A drawback for real-world applications is that kernels must be positive semi-definitive, therefore excluding some domain-specific similarity functions.

Other bag-based methods have addressed MIL by representing each bag by (dis)similarities to a set of prototypes $\mathcal{R} = \{R_1, \ldots, R_M\}$ in a so-called dissimilarity space~\cite{pekalska2005dissimilarity}. Therefore, each bag is represented by a single feature vector $ \mathbf{d}$$(B_i, \mathcal{R})= [d(B_i, R_1), \ldots, d(B_i, R_M)]$, where $d$ is a (dis)similarity measure. In this space, any supervised classifier can be used. In MILES~\cite{chen2006miles}, all training instances are used as prototypes, creating a very high-dimensional representation. A sparse classifier is used to select the most discriminative similarities, and therefore, instances. In a bag-of-words approach, prototypes are ``words'', or clusters of instances, and the dissimilarity measure between a bag and a word is the number of instances, belonging to that cluster.

Using bags as prototypes~\cite{tax2011bag,cheplygina2012does} reduces the dimensionality, and therefore, the possibility of overtraining. In this paper, we use all training bags as prototypes, i.e., $\mathcal{R} = \mathcal{X}_{tr}$, but prototype selection can be further used to reduce $|\mathcal{R}|$. The advantage of such methods is that more can be gained from the training data than in nearest neighbor approaches, and that there are no restrictions on the bag similarity function~\cite{pkekalska2006non}. In this paper, for example, we use two definitions of $d$ that are not necessarily metric: the average minimum instance distance (\ref{eq:meanmin}), and the earth mover's distance (EMD)~\cite{rubner2000earth}, defined in (\ref{eq:emd}). Herein, the instance dissimilarity $d$ is the squared Euclidean distance.

\begin{equation}~\label{eq:meanmin}
d_{meanmin}(B_i, B_j)  =  \frac{1}{n_i}\sum_{k=1}^{n_i} \min_{l} d(\mathbf{x}_{ik}, \mathbf{x}_{jl})
\end{equation}

\begin{equation}\label{eq:emd}
d_{\text{EMD}}(B_i,B_j) = \sum_{\mathbf{x}_{k} \in B_i, \mathbf{x}_{l} \in B_j}f(\mathbf{x}_{k},\mathbf{x}_{l}) d(\mathbf{x}_{k},\mathbf{x}_{l})
\end{equation}
where $f(\mathbf{x}_{k},\mathbf{x}_{l})$ is the flow that minimizes the overall distance, and that is subject to constraints that ensure that the only available amounts of ``earth'' (instances of $B_i$) are transported into available ``holes'' (instances of $B_j$), and that all of the instances are indeed transported: $f(\mathbf{x}_{k}, \mathbf{x}_{l}) \geq 0$, $\sum_{\mathbf{x}_{k} \in B_i} f(\mathbf{x}_{k}, \mathbf{x}_{l}) \leq 1/n_j$, $\sum_{\mathbf{x}_{l} \in B_j} f(\mathbf{x}_{k}, \mathbf{x}_{l}) \leq 1/n_i$ and $\sum_{\mathbf{x}_{k}\in B_i, \mathbf{x}_{l} \in B_j} f(\mathbf{x}_{k},\mathbf{x}_{l}) = 1$.


\section{Experiments}\label{sec:experiments}

\begin{figure}[ht]
 \centering
  \includegraphics[width=0.47\columnwidth]{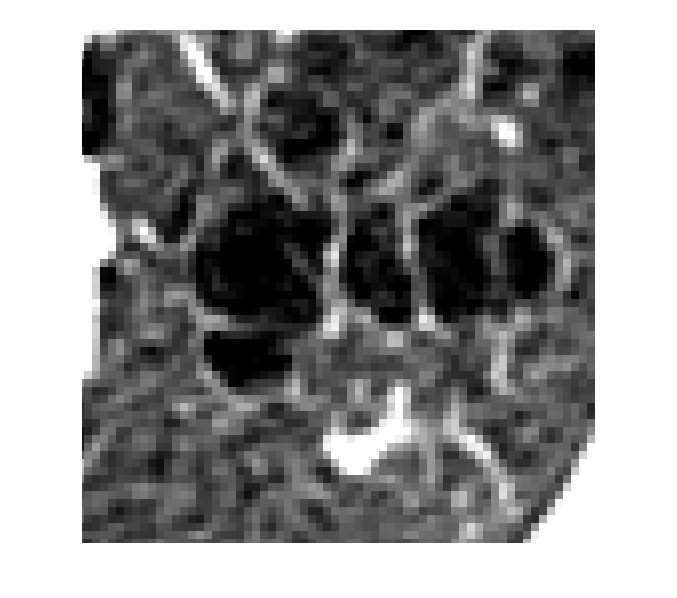}
  \includegraphics[width=0.47\columnwidth]{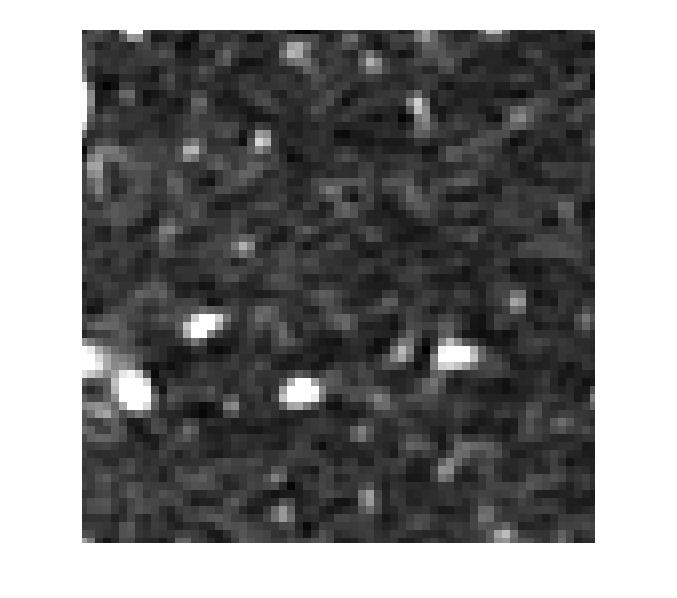}

 \caption[]{Examples of patches containing centrilobular emphysema (left), characterized by black holes within the lung tissue, and healthy tissue (right). Both images are approximately 1.5 times the size of the ROIs used for classification and the intensity values have been rescaled to facilitate viewing.}
 \label{fig:images}
\end{figure}

We use the dataset from~\cite{sorensen2012texture}, which describes how CT lung images from the Danish Lung Cancer Screening Trial~\cite{pedersen2009danish} have been processed. Parts of such images highlighting healthy and emphysemous lung tissue, are shown in Figure~\ref{fig:images}.

The dataset consists of three parts: training set $\mathcal{X}_{tr}$, validation set $\mathcal{X}_{val}$ and test set $\mathcal{X}_{te}$. Originally, each of these parts consists of 100 COPD (positive) and 100 healthy (negative) images. In previous work~\cite{sorensen2012texture}, a subset of the training data with 31 COPD and 31 healthy images was selected to improve the class separability in the training set. We therefore refer to the full training data as $\mathcal{X}_{tr}$ and to the subsampled training data as $\mathcal{X}_{sub}$.

Each image is represented by 50 ROIs, sampled at random locations within the lungs. Each ROI is described by histograms of responses of 8 filters at 7 scales, which aim to capture the texture of the image. The filters used the following: Gaussian, gradient magnitude, Laplacian of Gaussian, first, second and third eigenvalue of the Hessian, Gaussian curvature and eigen magnitude. The scales range from 0.6 to 4.8 mm. The responses of each filter at each scale are stored in a histogram with 41 bins. This approach creates a 2296-dimensional feature vector for each ROI. In~\cite{sorensen2012texture}, the validation set was used to select the most appropriate filters and scales. Because these features are selected for a particular classifier only (SimpleMIL with nearest neighbor classifier), we use the full feature set here for all the classifiers.

The evaluated classifiers are available from the MIL toolbox~\cite{MIL2011} and PRTools~\cite{prtools}. We evaluate the following selection:

\begin{itemize}

\item SimpleMIL with a logistic (regularization parameter $C \in \{0.01, 0.1, 1,10\}$) and nearest neighbor ($k \in \{25, 35, 45\}$) classifiers. We consider both noisy-or and average fusion rules.

\item EM-DD with 10\% of instances used for initialization

\item miSVM with a polynomial kernel, where $p \in \{1, 2\}$ is the degree of the polynomial and $C \in \{0.01, 0.1, 1, 10\}$ is a regularization parameter. We consider both noisy-or and average fusion rules.

\item MILBoost with 100 reweighting rounds

\item Citation $k$-NN, $k_R \in \{1, 5, 10\}$, $k_C \in \{1, 5, 10\}$

\item Averaging the instances (mean-inst), and minimum and maximum feature values for each bag (extremes) with an SVM, $p \in \{1, 2\}$, $C=\{0.01, 0.1, 1, 10\}$.

\item Bag-of-words (BoW) with \{50, 100, 200\} words and an SVM, $p \in \{1, 2\}$, $C \in \{0.01, 0.1, 1, 10\}$

\item MILES with a polynomial kernel, $p \in \{1, 2\}$, $C \in \{0.01, 0.1, 1, 10\}$

\item Bag dissimilarities (meanmin and emd) with $k$-NN, $k \in \{1, 5, 10\}$, and in the dissimilarity space with an SVM, $p \in {1, 2}$, $C=\{0.01, 0.1, 1, 10\}$

\end{itemize}

We perform evaluation in three ways. First, each classifier (with different parameter settings) is trained on $\mathcal{X}_{sub}$ and $\mathcal{X}_{tr}$, depending on the experiment. Each classifier is then evaluated on $\mathcal{X}_{val}$. The evaluation metric is the area under the receiver-operating characteristic curve, or AUC. We report the best of these performances on $\mathcal{X}_{val}$, and select the corresponding parameters. We then report the performance of this classifier with the best parameters on an independent test set $\mathcal{X}_{te}$. The difference in AUC on $\mathcal{X}_{val}$ and $\mathcal{X}_{te}$ is an indicator of overtraining, i.e., fitting the parameters too well to the validation set. Lastly, we randomly select half of the bags in $\mathcal{X}_{sub}$ or $\mathcal{X}_{tr}$, 10 times. For each subsample, we train a classifier, select parameters using $\mathcal{X}_{val}$, and evaluate on $\mathcal{X}_{te}$. The average and the standard deviations of the 10 performances are reported. This result gives an indication of a situation where less training data is available, and of the variance in performance due to a different sampling of the data.

The performances are shown in Table~\ref{tab:res_simple}. For each training dataset, we compare the performances per column. The best performance and performances not significantly worse than best, are shown in bold. We test for significant differences using the DeLong test for ROC curves~\cite{delong1988comparing} for the performances on $\mathcal{X}_{val}$ and $\mathcal{X}_{te}$, and using a dependent t-test for the 10 cross-validation performances, both at a significance level of 0.05. A few results are not reported. For EM-DD, time requirements were too high for both datasets. For miSVM, the instance kernel matrix for $\mathcal{X}_{tr}$ was too large to fit in memory.

\begin{table}[ht]

\caption{AUC performances ($\times 100$) of MIL classifiers, trained on $\mathcal{X}_{sub}$ (top) and $\mathcal{X}_{tr}$ (bottom). From left to right: best parameters on $\mathcal{X}_{val}$,  same parameters on $\mathcal{X}_{te}$, mean(std) when subsampling $\mathcal{X}_{sub}$ or $\mathcal{X}_{tr}$ to 50\% 10 times}

\begin{center}
\begin{tabular}{l l l l}



 & \multicolumn{3}{c}{Trained on $\mathcal{X}_{sub}$} \\
  Classifier &    AUC $\mathcal{X}_{val}$ & AUC $\mathcal{X}_{te}$ & 10x AUC $\mathcal{X}_{te}$ \\

\hline

Simple logistic noisy & 50.0 & 50.0 & 50.2 ( 0.7)\\
Simple logistic avg   & {\bf 71.9 }& {\bf 70.5 }& 67.9 ( 1.3)\\
Simple $k$-NN noisy     & 61.0 & 65.9 & 63.7 ( 2.3)\\
Simple $k$-NN avg       & 67.0 & 67.8 & 66.0 ( 1.5)\\
miSVM noisy           & {\bf 69.7 }& 65.4 & 62.0 ( 3.1)\\
miSVM avg             & {\bf 74.5 }& {\bf 71.7 }& 69.4 ( 1.5)\\
MILBoost 		           & 55.8 & 61.4 & 59.3 (10.2)\\
Citation $k$-NN         & 65.2 & 61.5 & 63.5 ( 1.5)\\
mean-inst SVM         & {\bf 74.0 }& {\bf 74.2 }& {\bf 72.3 ( 2.7)}\\
extremes SVM          & {\bf 70.8 }& 68.6 & 68.3 ( 2.7)\\
BoW SVM               & 50.0 & 50.0 & 50.0 ( 0.0)\\
MILES                 & 65.8 & 68.2 & 64.3 ( 4.2)\\
meanmin SVM           & 70.8 & {\bf 71.3 }& 69.6 ( 2.1)\\
meanmin $k$-NN          & 65.0 & {\bf 69.1 }& 65.7 ( 1.6)\\
emd SVM               & {\bf 73.7 }& {\bf 74.6 }& 69.3 ( 3.3)\\
emd $k$-NN              & 65.1 & 67.1 & 64.6 ( 1.8)\\

& & & \\

 & \multicolumn{3}{c}{Trained on $\mathcal{X}_{tr}$} \\

                    Classifier &    AUC $\mathcal{X}_{val}$ & AUC $\mathcal{X}_{te}$ & 10x AUC $\mathcal{X}_{te}$ \\
 \hline

Simple logistic noisy & 60.9 & 60.7 & 50.0 ( 0.0) \\
Simple logistic avg   & {\bf 73.5 }& {\bf 75.8 }& 72.3 ( 3.1)\\
Simple $k$-NN noisy     & 64.3 &  68.2 &  66.9 ( 2.1) \\
Simple $k$-NN avg       & 66.8 & 69.7 & 68.5 ( 0.8) \\
MILBoost & 54.6 & 54.3 & 62.3 ( 7.8) \\
Citation $k$-NN & 65.9 & 56.9 & 60.4 ( 2.4) \\
mean-inst SVM         & {\bf 77.2 }& {\bf 77.6 }& {\bf 76.5 ( 3.8)}\\
extremes SVM          & {\bf 73.1 }& 65.2 & 67.2 ( 1.2)\\
BoW SVM               & 50.0 & 50.0 & 50.0 ( 0.0) \\
MILES 								& 50.0 & 50.0 & 67.6 ( 2.5) \\
meanmin SVM           & {\bf 74.0 }& {\bf 75.4 }& 73.8 ( 2.6)\\
meanmin $k$-NN          & 59.0 & 53.5 & 53.5 ( 4.6) \\
emd SVM               & {\bf 74.2 }& 72.9 & {\bf 75.1 ( 2.7)}\\
emd $k$-NN              & 63.9 & 54.4 & 51.2 ( 4.3) \\

\end{tabular}
\end{center}\label{tab:res_simple}
\end{table}


\section{Discussion}\label{sec:results}

\subsection{Classifier Performance}

Across the different training datasets, we can see similar trends in the classifier performances. It is clear that some classifiers suffer from the high dimensionality. For example, the bag of words approach, which uses a mixture of Gaussians to estimate words in feature space, is not able to do so in 2296 dimensions. For BoW, MILBoost, and EM-DD which could not handle the dimensionality computationally, we performed additional experiments with the 287-dimensional feature set that resulted from a feature selection procedure used in~\cite{sorensen2012texture}. The results on $\mathcal{X}_{te}$ are 0.657 (BoW), 0.641 (EM-DD) and 0.551 (MILBoost), suggesting that these classifiers benefit from feature selection. It may of course generally be interesting to study feature selection for the other classifiers as well.

The full training dataset $\mathcal{X}_{tr}$ has two main differences with respect to $\mathcal{X}_{sub}$: higher class overlap, and more bags, and therefore instances in total. Several classifiers, such as SimpleMIL logistic, mean-inst, extremes, and SVM in the dissimilarity space, show increases in performances due to the higher sample size. On the other hand, MILES suffers from the increased sample size, because the dimensionality of the dissimilarity representation is equal to the number of instances. This also explains why MILES performs better when the training set is subsampled to 50\%.

The performances of many methods do not degrade very much when only 50\% of the bags are used for training. This suggests that the subsampled dataset is still representative for the whole data distribution, and each class can be described well with only a few samples. Furthermore, most classifiers do not suffer a lot from overtraining, as the difference in performance on $\mathcal{X}_{val}$ and on $\mathcal{X}_{te}$ is quite small. Notable exceptions are the $k$-NN classifiers trained on the full training set $\mathcal{X}_{tr}$, where the parameter $k$ is overfit to the validation set, causing lower performances on $\mathcal{X}_{te}$.


SimpleMIL performs quite well, especially when posterior probabilities of all instances are taken into account, as in the averaging fusion rule. Methods which assume a concept, such as EM-DD and miSVM, also perform reasonably, which suggests that there is a region in feature space with a high density of disease patches and low density of normal patches. However, the performances are lower than those of bag-based methods, suggesting that detecting the concept is not sufficient for the diagnosis of COPD. This is also supported by the fact that miSVM with the averaging rule outperforms miSVM with the noisy or rule, which shows that it is beneficial to take all instance classifications into account.

Methods with assumptions on bag level have the best performances, in particular, averaging all the instances in a bag is already able to separate the bag classes quite well. This suggests that negative instances in positive bags, and the negative instances in negative bags, do not originate from the same distribution. In other words, scans affected with COPD do not contain the same types of healthy patches, as healthy scans. The disease appears to be more diffuse, affecting a large part of the lung rather than small isolated regions.

For the bag-based methods, the mean-inst bag representation and the dissimilarity-based SVM perform particularly well. MILES suffers from the high dimensionality, but we expect that the performance would improve if instance selection techniques would be used. Another interesting observation is that the dissimilarity-based SVM significantly outperforms $k$-NN on the same dissimilarities. SVM is able to use the dissimilarities of the training set to create a more robust classifier, which is consistent with results in~\cite{sorensen2010image}, although slightly different dissimilarities are used there. We expect that further investigation into different bag dissimilarity measures could further improve these results.

Unfortunately, our results are not directly comparable to the dissimilarity-based approach of~\cite{sorensen2010image}, because an earlier version of the dataset was used. The results in~\cite{sorensen2012texture}, however, are obtained by training on $\mathcal{X}_{sub}$ and the performances can be compared. There, the best approach obtains an AUC of 0.713. Our results show superior performances when training on $\mathcal{X}_{sub}$, with an AUC 0.742 for mean-inst and 0.746 for $d_{\text{EMD}}$ in the dissimilarity space. However, these performances are not significantly better than the result in~\cite{sorensen2012texture}. Using $\mathcal{X}_{tr}$ further improves the results, for an AUC of 0.758 for SimpleMIL with a logistic classifier, 0.776 for mean-inst, and 0.754 for $d_{meanmin}$ in the dissimilarity space. The best approach using $\mathcal{X}_{tr}$ in ~\cite{sorensen2012texture} obtains an AUC of 0.690, and our performances are significantly better according to the DeLong test. 


Furthermore, we examined the output of the best-performing classifiers to see which images still get misclassified. Rather than only looking at the positive label for COPD, we now use the COPD stages~\cite{rabe2007global}, from mild (I) to moderate (III). The results show that most of the confusion is between the healthy scans, and stage I scans, which supports our intuition about where the class overlap is largest. Because the classifiers differ in some of their errors, it may be of interest to combine their decisions.


\subsection{Concept Region}



\begin{figure}[ht]
 \begin{center}
  \includegraphics[width=0.75\columnwidth]{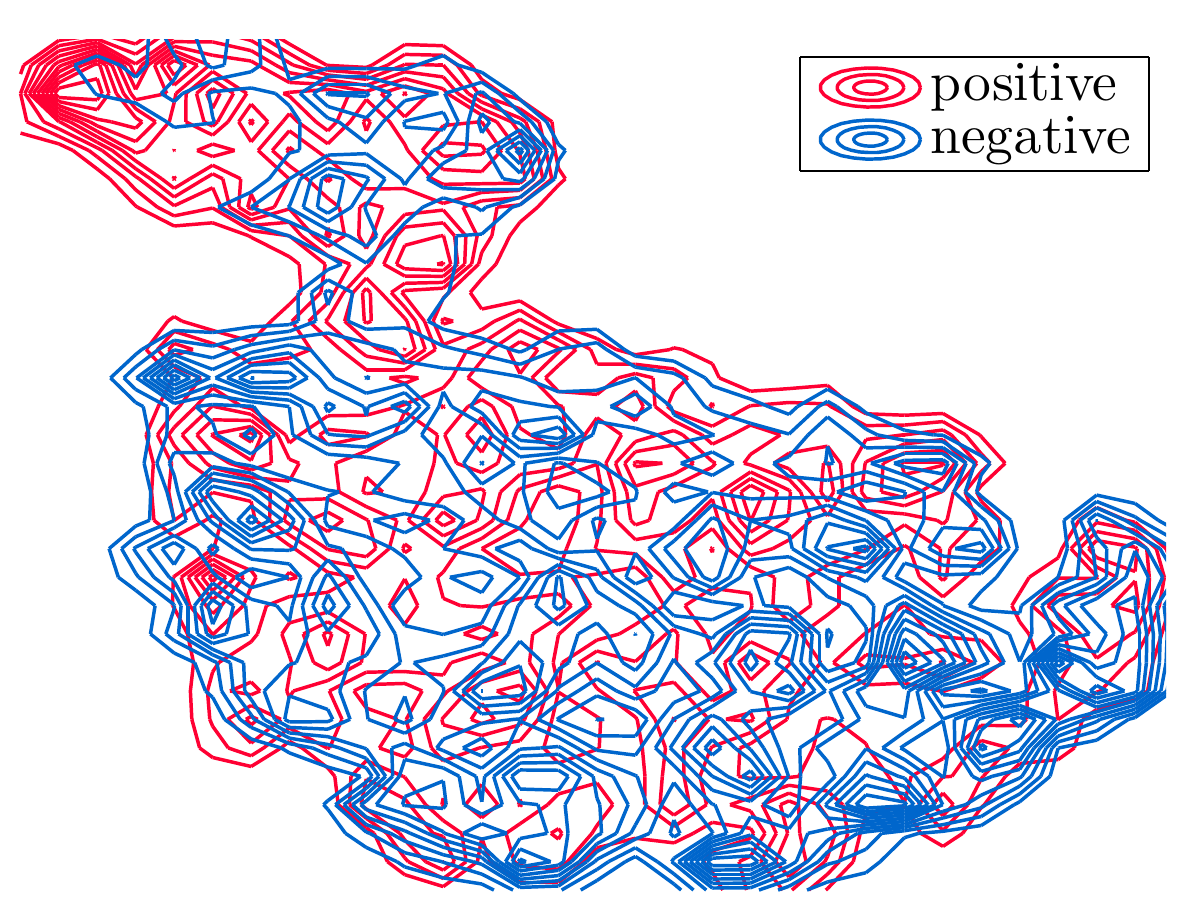}
 \end{center}
 \caption[]{Density contours of t-SNE projection of instances of $\mathcal{X}_{sub}$}
 \label{fig:tsne}
\end{figure}

\begin{figure}[ht]
 \centering
  \includegraphics[width=0.75\columnwidth]{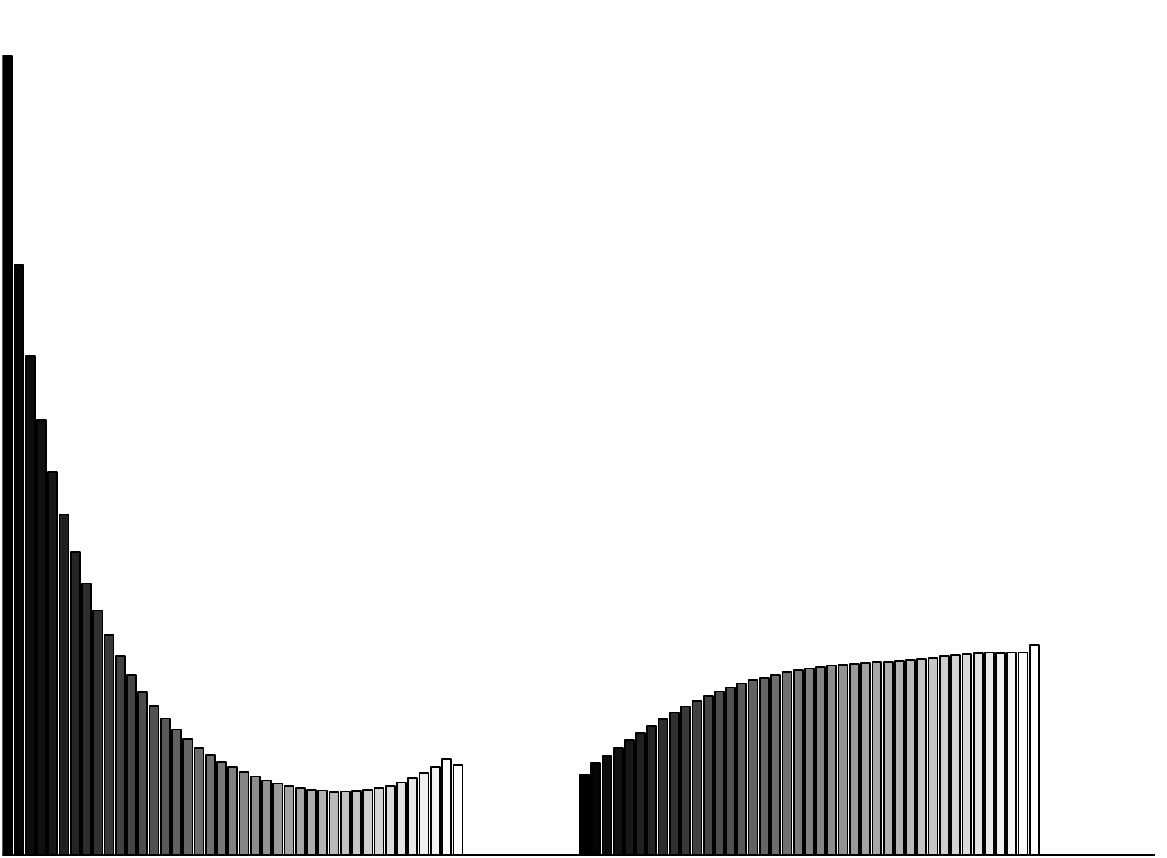}

 \caption[]{Histograms of Gaussian filter responses at the scale 0.6, for the averaged instances in the two clusters found in the t-SNE plot.}
 \label{fig:hist}
\end{figure}

In order to better understand the classifier performances, we examine a 2D projection of the instances, obtained by t-distributed stochastic neighbor embedding (t-SNE)~\cite{van2008visualizing} (Figure~\ref{fig:tsne}). We see two clusters of instances, a smaller cluster in the top left and a larger cluster. In the small cluster the density of instances from positive bags is clearly higher, which suggests that part of it could be a concept region. To investigate whether these patches display emphysema, we examined the intensity histograms of the Gaussian filters at the smallest scale. As emphysema results in darker patches, we would expect patches with emphysema to have intensity histograms skewed to the left. This is exactly what we find when averaging all the instances per cluster and plotting the two corresponding histograms in Figure~\ref{fig:hist}.

Visual inspection of patches from both the small cluster and from the lower right part of the big cluster in Figure~\ref{fig:tsne} confirmed the tendency we saw in the average Gaussian filter response histograms in Figure~\ref{fig:hist}. The patches in the small cluster were generally affected by emphysema whereas the patches in the lower part of the big cluster showed no or only faint signs of emphysema.

It is important to note that the dataset mainly contains mild to moderate COPD patients, and no patients with very severe emphysema. We expect that if this was the case, the concept or concepts would be more pronounced.

\subsection{Interpretability}

Next to the classifier performances, it is important to consider how these classifiers would be used in a medical setting. Despite slightly lower performances, instance-based methods are of interest because of their ability to provide instance labels for the ROIs. An expert could then inspect the instance labels in different regions of the lungs, allowing for better diagnosis or treatment planning. The instance labels, however, should be used with caution. Specialized MIL (i.e., except SimpleMIL) methods are trained to classify bags correctly, not instances, and the best bag classifier is not necessarily the best instance classifier~\cite{tragante2011instance}. Therefore, correct instance labels would be sacrificed for the greater good of correct bag labels.

Although bag-based methods perform better, their interpretability may be more difficult. For example, the average
histograms (as in mean-inst) separate the classes very well, but this method can not provide information on how the
affected tissue is distributed within the lungs, which could be important for determining the best treatment as well as for monitoring disease progression and therapy effect.

Dissimilarity-based methods provide more opportunities in terms of interpretability compared to mean-inst or extremes. For these methods, we can investigate which prototypes, i.e. CT images, patch clusters or individual patches, correspond to typical healthy or COPD cases. By using linear classifiers in the dissimilarity space, the diagnosis would be explained in terms of a linear combination of dissimilarities to such prototypes.

\section{Conclusions}\label{sec:conclusion}

We have studied the possibility of classifying COPD by means of various classical and more recent MIL approaches. The study revealed that MIL offers classification methods for this problem that are potentially better than the techniques previously proposed.  The diversity of methods also enabled us to reason about the nature of COPD as a MIL problem. Although we found a concept region with patches showing typical disease patterns, considering the whole distribution of instances for bag classification improved the results. The best performing method is an SVM with a kernel based on the average instance per bag. This method obtains an AUC of 0.742 which is higher (but not significantly) than the previous best performance of 0.713 on the same dataset. By using the full training data we achieve a significantly higher AUC of 0.776.

\IEEEtriggeratref{20}

\bibliographystyle{IEEEtran}
\bibliography{refs}

\end{document}